\documentclass{article}
\usepackage{iclr2027_conference,times}
\usepackage[T1]{fontenc}
\usepackage{amsmath,amssymb}
\usepackage{booktabs,tabularx,array}
\usepackage{graphicx,xcolor}
\usepackage{hyperref,url}
\AtBeginDocument{}
\hypersetup{breaklinks=true,pdftitle={Graph-Conditioned On-Policy Agent Distillation from Off-the-Shelf Teachers},pdfauthor={Xiaohan Yi, Wen Luo, Yani Huang, Junfeng Zhan, Asher Qin, Peilin Zhao, Xi Xiao}}

\newcommand{\method}{GC-OPD}
\newcommand{\meanstd}[2]{\mbox{#1{\fontsize{5}{5.5}\selectfont\color{gray}$\pm#2$}}}

\title{Graph-Conditioned On-Policy Agent\\Distillation from Off-the-Shelf Teachers}
\author{%
Xiaohan Yi\textsuperscript{1,2,*},
Wen Luo\textsuperscript{1,3,*},
Yani Huang\textsuperscript{1,*},
Junfeng Zhan\textsuperscript{1},\\[1pt]
\bfseries Asher Qin\textsuperscript{1},
Peilin Zhao\textsuperscript{4},
Xi Xiao\textsuperscript{2,\textdagger}\\[2pt]
{\small\textsuperscript{1}Yuanbao Team, Tencent\quad\textsuperscript{2}Tsinghua University}\\
{\small\textsuperscript{3}Huazhong University of Science and Technology}\\
{\small\textsuperscript{4}School of Artificial Intelligence, Shanghai Jiao Tong University}\\[2pt]
{\small\textsuperscript{*}Equal contribution.\quad\textsuperscript{\textdagger}Corresponding author.}\\[1pt]
{\small\href{mailto:yxh24@mails.tsinghua.edu.cn}{\texttt{yxh24@mails.tsinghua.edu.cn}}\quad\href{mailto:xiaox@sz.tsinghua.edu.cn}{\texttt{xiaox@sz.tsinghua.edu.cn}}}
}

\usepackage{etoolbox}
\makeatletter
\patchcmd{\@maketitle}{\vskip 0.3in minus 0.1in}{\vskip 0.05in minus 0.02in}{}{\PackageError{gcopd}{Title spacing patch failed}{Check the ICLR style}}
\patchcmd{\@maketitle}{\rule{\z@}{24pt}}{\rule{\z@}{16pt}}{}{\PackageError{gcopd}{Author spacing patch failed}{Check the ICLR style}}
\makeatother
\DeclareMathSizes{8}{9}{7}{5}
\DeclareMathSizes{6.5}{9}{7}{5}
\DeclareMathSizes{7.5}{9}{7}{5}

\fancypagestyle{yuanbaofirstpage}{%
  \lhead{\raisebox{-1.2pt}{\includegraphics[height=10pt]{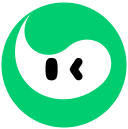}}\ Tencent Yuanbao}%
}
\iclrfinalcopy
\begin{document}
\vspace*{-24pt}
\maketitle
\lhead{Preprint}
\thispagestyle{yuanbaofirstpage}
\begin{abstract}
On-policy distillation (OPD) trains compact language agents with teacher
feedback on student-generated trajectories. In multi-turn tasks, compounding
errors can move students beyond the teacher's effective supervision.
We introduce Graph-Conditioned On-Policy Agent Distillation (GC-OPD),
which enriches an off-the-shelf teacher's scoring context with execution
evidence. A graph indexes repeated teacher executions by shared states
while preserving complete successful and failed histories. After each
student episode, GC-OPD retrieves current-state references or historical
alternatives and combines them with student hindsight to score the
original thought--action tokens. Using the same original teachers,
GC-OPD improves mean success over vanilla OPD from 24.70\% to 48.78\%
on ScienceWorld (4B student), from 53.36\% to 85.26\% on ALFWorld Unseen,
and from 29.10\% to 37.65\% on WebShop. At matched student sizes, it also
achieves higher mean success than every evaluated OPD baseline using
GRPO-trained teachers on ScienceWorld and ALFWorld; the strongest such
ScienceWorld 4B baseline reaches 46.66\%. GC-OPD requires no task-specific
teacher optimization.
\end{abstract}

\begin{center}
{\small Code: \url{https://github.com/hanyi2021/GC_OPD}\par}
\vspace{2pt}
\begin{minipage}{\linewidth}
\centering
\includegraphics[width=\linewidth]{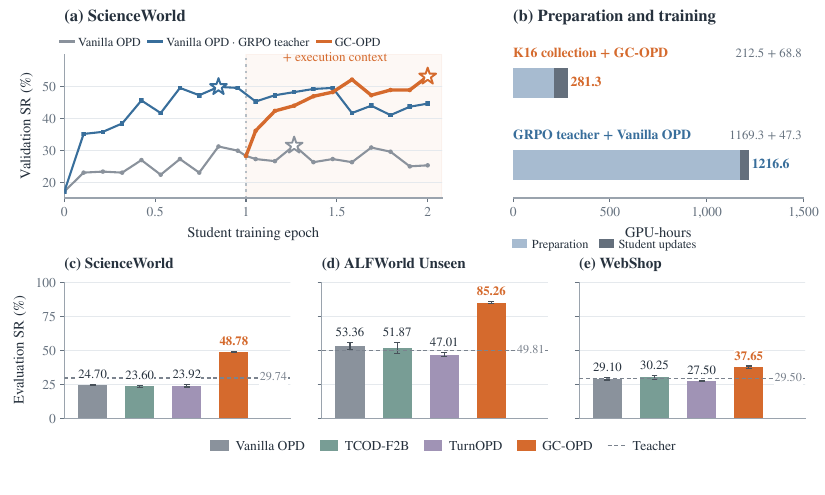}
\refstepcounter{figure}\label{fig:homepage_teaser}
\par\vspace{2pt}
\begin{minipage}{\linewidth}\footnotesize
\textbf{Figure~\thefigure: Distilling agents from off-the-shelf teachers.}
(a) ScienceWorld 4B validation curves (305 tasks, one seed per checkpoint); stars mark selected checkpoints. (b) Preparation and training for a 1.7B student; costs sum measured preparation and student-training stages using 64 and eight H20 GPUs, respectively. (c--e) Distillation from original teachers: student success, mean$\pm$SD over four inference seeds; dashed lines show teacher success. Cost accounting is in Appendix~\ref{app:compute_details}.
\end{minipage}
\end{minipage}
\end{center}

\section{Introduction}
\label{sec:introduction}
On-policy distillation (OPD) trains compact language agents with teacher
feedback on responses sampled from the student's own policy
\citep{gu2024minillm,agarwal2024policy,tmlopd}.
In multi-turn environments, early errors change the states encountered
later: the student can drift beyond the teacher's effective support,
degrading supervision as interaction continues \citep{ross2011dagger,wang2026tcod}.
Existing methods address this problem through the interaction process.
TCOD-F2B gradually increases student rollout depth; TCOD-B2F replays
successful prefixes before handing control to the student
\citep{wang2026tcod}. FTB intervenes at high-disagreement decisions and
uses subsequent student continuations to assess teacher-proposed alternatives
\citep{chen2026futurebridge}. These approaches highlight a central challenge:
\emph{maintaining useful supervision along the executions that students
actually produce, including after they deviate.}

Teacher preparation presents a second challenge. Some agent-distillation
studies use task-specialized or reinforcement-learning-trained teachers
\citep{zhou2026turnopd,chen2026futurebridge}. Preparing a large teacher
adds optimization and environment-interaction costs. It also requires
memory for gradients, optimizer states, and backward activations beyond
the weights needed for inference \citep{rajbhandari2020zero}.
An off-the-shelf teacher avoids this additional optimization stage,
but its task performance may initially be limited. The challenge is to
use its available experience to train students competitive with those
distilled from task-trained teachers.

Repeated execution exposes capabilities that a single attempt misses.
In a fixed ScienceWorld evaluation bank, the original Qwen3-32B teacher
\citep{yang2025qwen3} achieves 29.60\% pass@1 and 66.83\% pass@16
(Appendix Table~\ref{tab:teacher_pass16}).
These attempts record both successful solutions and the consequences of
unsuccessful choices. \emph{Can this execution experience improve the
teacher's supervision without first improving its parameters?}
We investigate this question using repeated executions collected on
training tasks, separate from the evaluation bank.

Two design challenges arise. \textbf{Evidence selection:} records span
many states and outcomes, so the teacher needs relevant experience without
receiving the entire library or losing the history that explains an outcome.
\textbf{Trajectory divergence:} a task may have recorded solutions even
when no eligible successful record matches the student's current state. Earlier student states
may still connect to successful alternatives; those records must be
identified and distinguished from the student's actual continuation.

We introduce \emph{Graph-Conditioned On-Policy Agent Distillation}
(\method). Shared state locators connect visits across recorded executions,
while source identities preserve their complete histories and outcomes.
These graph associations guide retrieval at the student's current state
or, when eligible successful support is absent, earlier visited states.
Remaining interaction cost ranks the associated successful continuations;
retained records are provided in full so the teacher can account for the
observations and preparations preceding their outcomes.
After each episode, the fixed teacher conditions on the selected execution
records and student hindsight to score the student's original thought--action
tokens. Only the student is updated.

Our contributions are threefold:
\begin{itemize}
\setlength{\itemsep}{1pt}
\item We introduce GC-OPD, which turns recorded interactions into
training-time supervision from an off-the-shelf teacher without
additional teacher optimization.
\item We develop source-preserving state and history retrieval, linking
student decisions to complete records and exposing historical alternatives
when no current-state successful reference is available.
\item At matched student sizes, GC-OPD with off-the-shelf teachers achieves
higher mean success than all evaluated GRPO-teacher OPD baselines on
ScienceWorld and ALFWorld. ScienceWorld gains over the best such baselines
are 7.48 and 2.12 percentage points for 1.7B and 4B students, with fewer
interactions. WebShop also improves over vanilla OPD; ablations examine
which execution evidence contributes to supervision.
\end{itemize}

\section{Related Work}
\label{sec:related_work}
\subsection{On-Policy Distillation for Large Language Models}
Knowledge distillation transfers teacher predictions to a smaller model
\citep{hinton2015distilling}. For autoregressive models, teacher-generated
training sequences can differ from the contexts encountered by the student.
On-policy methods address this mismatch through supervision on student
samples, following the principle of learning under the learner's own state
distribution \citep{ross2011dagger}. MiniLLM and generalized knowledge
distillation develop distribution-matching objectives for this setting
\citep{gu2024minillm,agarwal2024policy}.
Teacher context provides another source of supervision: OPCD studies
context-conditioned knowledge transfer \citep{ye2026opcd}; OPID and SEED
extract hindsight skills to rescore original responses
\citep{yang2026opid,wu2026seed}; Skill-SD retrieves task-local skills
\citep{wang2026skillsd}. PAST combines complete-response privilege with
teacher adaptation \citep{feng2026past}.
GC-OPD builds on context-conditioned OPD by selecting complete external
executions through the student's current and historical states.

\subsection{Credit Assignment in Long-Horizon Agent Tasks}
Delayed outcomes provide limited information about which earlier decisions
enabled success or caused failure. Policy-gradient methods connect returns
to local updates \citep{schulman2017ppo}; hindsight experience replay reuses
failed experience through goal relabeling \citep{andrychowicz2017hindsight},
and Reflexion converts feedback into reusable verbal memory
\citep{shinn2023reflexion}. Recent agent-distillation methods provide more
localized supervision. TCOD controls student interaction depth with a
curriculum \citep{wang2026tcod}; TurnOPD allocates rollout depth and loss
across turns \citep{zhou2026turnopd}; ATOD adjusts the balance of distillation
and reinforcement learning and reweights turn-level signals
\citep{tan2026atod}. FTB assesses local teacher interventions using subsequent
student continuations \citep{chen2026futurebridge}. AgentOPSD aggregates
privileged token-level evidence into turn-level credit
\citep{wang2026agentopsd}. Graph-based methods
use relations among executions to structure credit assignment
\citep{cheng2026graphgpo,wang2026g2po,gan2026tigpo}; DART-SD uses
teacher-execution graphs to generate recovery responses for masked
supervised learning \citep{xu2026dartsd}.
GC-OPD uses state-associated records and student hindsight to inform
feedback on the student's original decisions, including after divergence.

\section{Graph-conditioned on-policy agent distillation}
\label{sec:method}
\subsection{Preliminaries}
\label{sec:preliminaries}
For task $x$, a language agent interacts with an environment over multiple
decisions. The student policy $p_\theta$ receives public context $h_t$:
the task, current observation, environment-provided action information,
and a bounded observation--action history.
It samples a thought--action response $y_t=(y_{t,1},\ldots,y_{t,|y_t|})$
\citep{yao2023react}; the environment executes its parsed action and returns
an observation and feedback. Let $\tau$ denote the completed student episode,
including actions, observations, feedback, and outcome.

In vanilla OPD, a fixed teacher evaluates each student-sampled
token given the same history $h_t$ and response prefix $y_{t,<i}$
\citep{agarwal2024policy,tmlopd}. The formulation used here supplies the
teacher--student log-probability difference as token-level feedback:
\begin{equation}
\Delta_{t,i}
=\log q_{\mathrm T}(y_{t,i}\mid h_t,y_{t,<i})
-\log p_\theta(y_{t,i}\mid h_t,y_{t,<i}).
\label{eq:ordinary_opd_feedback}
\end{equation}
Only the student is updated. GC-OPD retains this interface and adds
execution records to the teacher's scoring context; the student continues
to use its original decision-time context.

\begin{figure}[!tbp]
\centering
\includegraphics[width=\linewidth]{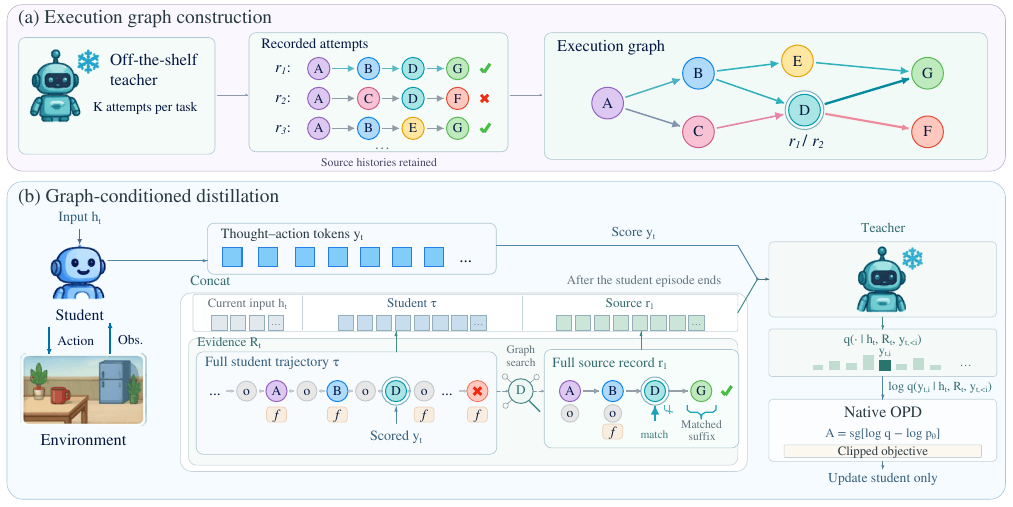}
\caption{\textbf{GC-OPD overview.} (a) Shared state nodes associate visits across recorded executions for reference retrieval, while source identities preserve complete histories. (b) The selected complete records and student hindsight condition the fixed teacher's scoring of the original student tokens; only the student is updated. State and history retrieval is specified in Section~\ref{sec:retrieval}, with the full procedure in Algorithm~\ref{alg:gcopd} (Appendix~\ref{app:retrieval_algorithm}).}
\label{fig:method}
\end{figure}

\paragraph{Two-stage training overview.}
GC-OPD first trains the student with vanilla OPD, whose teacher scores
responses using the original decision-time context $h_t$. It then continues
training the warmed-up student with graph-conditioned feedback: the teacher
additionally receives student hindsight and selected execution records.
Both stages count toward the student-training budget, and the teacher
remains frozen throughout student training.
For each task $x$, a generator $q_{\mathrm{gen}}$ collects a fixed offline
library $\mathcal D_x$; the main configuration uses the same off-the-shelf
model for generation and scoring. In the GC stage, the student completes
each episode before retrieval and scoring. \textbf{Execution evidence is
supplied only to the teacher.} Figure~\ref{fig:method} summarizes this stage.

\subsection{Source-preserving execution graph}
\label{sec:graph_construction}
Consider A--B--D--G and A--C--D--F in Figure~\ref{fig:method}.
Their distinct visits to D share a locator, associating the executions
for reference retrieval. Their prefixes can nevertheless contain different
observations and preparations: opening and closing a drawer can restore
its physical configuration while revealing its contents. D therefore
provides a comparison anchor; the complete source history supplies the
context needed to interpret its continuation and outcome.
A shared locator alone does not make the two prefixes and suffixes
interchangeable.

For a source $r$ with $L_r$ decisions, visit $v=(r,j)$ identifies the state
before decision $j$, for $j\in\{0,\ldots,L_r\}$; $j=L_r$ is terminal.
A training-only state descriptor $\phi(x,s)$ and interaction metadata
$\chi$ define a locator $z=\zeta(\phi(x,s),\chi)$
\citep{vapnik2009privileged,chen2019cheating}. Continuous quantities, such as
temperature in ScienceWorld, are discretized before matching. Unreliable captures receive
$z=\bot$ and cannot match. Appendix~\ref{app:state_descriptors} specifies
the environment-dependent descriptors and matching rules.

The source-visit graph $H_x=(\mathcal W_x,\mathcal F_x)$ contains these
visits and consecutive temporal edges $((r,j),(r,j+1))$. For all reliably
located visits $\mathcal V_x\subseteq\mathcal W_x$, projection
$\pi_x(v)=z(v)$ induces
\begin{equation}
\begin{aligned}
G_x&=(V_x,E_x),\\
V_x&=\{\pi_x(v):v\in\mathcal V_x\},\\
E_x&=\{(\pi_x(u),\pi_x(v)):(u,v)\in\mathcal F_x,\ u,v\in\mathcal V_x\}.
\end{aligned}
\label{eq:graph_nodes}
\end{equation}
Each shared node $z$ in $G_x$ links the source visits assigned to it:
\begin{equation}
\mathcal I_x(z)=\{v\in\mathcal V_x:\pi_x(v)=z\},
\qquad \operatorname{src}(r,j)=r.
\label{eq:node_visit_index}
\end{equation}
Here $\mathcal I_x(z)=\varnothing$ when $z\notin V_x$, including $z=\bot$.
A student locator queries $\mathcal I_x$ to identify source visits; the
source map $\operatorname{src}$ recovers their complete records from
$\mathcal D_x$. Temporal edges retain recorded transitions. Although
$G_x$ may contain cycles or self-loops, chronological positions in $H_x$
keep each recorded suffix finite (Appendix~\ref{app:graph_properties}).

Graph augmentation adds executed planner or oracle records with their
origin and observed outcomes under the same source-validity interface.
These additional executions, which may fail or exceed the student's
horizon, expand the library without changing the student objective
(Appendix~\ref{app:source_collection}).

\subsection{State- and history-conditioned retrieval}
\label{sec:retrieval}
\paragraph{Current-state evidence.}
A task can have successful source records without a matching visit at
the student's current decision.
For decision $t$ in completed student episode $\tau$, let $z_t$ be its
locator. We query the shared-node associations in $G_x$ and filter the
returned visits for eligible successful continuations:
\begin{equation}
\mathcal A_t^+=\{(r,j)\in\mathcal I_x(z_t):b^+(r,j)=1\}.
\label{eq:recorded_support}
\end{equation}
Here $b^+$ checks the source outcome,
recorded suffix continuity, and entry validity.
When $\mathcal A_t^+$ is nonempty, its visits compete with the student's
own continuation $(\tau,t)$ if the student succeeded. Denote this
candidate set by $\mathcal C_t^+$ and select
\begin{equation}
(r_t^*,j_t^*)\in\arg\min_{(r,j)\in\mathcal C_t^+}d(r,j),\qquad
 d(r,j)=\sum_{\ell=j}^{L_r-1}c_\ell^r.
\label{eq:suffix}
\end{equation}
Here $c_\ell^r$ is the interaction cost, with counting conventions in
Appendix~\ref{app:retrieval_algorithm}.
External suffixes follow their source's temporal edges in $H_x$.
Ranking from the matched visit prevents earlier source detours from
dominating the choice. At D, D--G determines the remaining cost, while
the complete record A--B--D--G preserves the observations and preparations
needed to interpret that continuation. Thus
$\operatorname{External}(v)=\{\operatorname{src}(v)\}$ for an external
winner, and is empty if the student's own continuation wins.

\paragraph{Historical evidence.}
When $\mathcal A_t^+=\varnothing$, earlier student locators query the
same graph associations. We choose the most recent anchor with eligible
successful support:
\begin{equation}
\alpha_t=\max\{u\leq t:\mathcal A_u^+\ne\varnothing\},\qquad
\max\varnothing:=\bot.
\label{eq:historical_anchor}
\end{equation}
For a student that visits D and then reaches an unmatched X, the earlier
D locates A--B--D--G as evidence for comparing the two continuations.
The anchor identifies an earlier point of comparison; it neither rolls
back the student nor inserts source actions into its execution.
Without an anchor, an eligible same-task successful record may provide an
explicitly unaligned fallback.

\paragraph{Reference composition.}
When no current-state successful reference is available, matched failed
records are found independently through the same index $\mathcal I_x$
at the latest shared student-history position,
so their anchors can differ from successful references. Their observed consequences remain useful evidence without
labelling every action in a failed episode as wrong.
Candidate availability and retention remain distinct:
$\operatorname{Retain}_{\mathrm{env}}$ applies the environment's assembly
rules using the candidates, current-state match, and student outcome.
Exact gates, fallbacks, cost counting, and tie-breaking appear in
Appendix~\ref{app:retrieval_algorithm}.

For each scored decision, the teacher context contains the complete student
execution and, when retained, at most one successful and one failed complete source. Each retains actions,
observations, feedback, and outcome; source identity, verification,
and association labels relate the records to the scored decision.
Historical thoughts are excluded. Student hindsight remains available
even without an external successful source. Tasks lacking successful
source records remain in the training set.

\subsection{Execution-conditioned on-policy distillation}
\label{sec:execution_distillation}
For a task batch $B$, freeze the rollout policy $p_{\mathrm{old}}\leftarrow
p_\theta$ and complete student episodes before assembling evidence:
\begin{equation}
\mathcal B_\tau\leftarrow\operatorname{Rollout}(p_{\mathrm{old}},B),
\qquad R_t\leftarrow\operatorname{Render}(\tau,t,\mathcal S_t).
\label{eq:rollout_render}
\end{equation}
Here $\mathcal S_t$ contains the complete records retained by the rules in
Section~\ref{sec:retrieval}; matched visits recover their records through
$\operatorname{src}$. Graph associations guide reference selection,
while $R_t$ renders full source histories alongside the student execution.
The teacher scores the same
response IDs that the student scores under its original public context
$h_t$; no missing action is inserted or substituted. Action-token scoring
therefore retains the student's original thought prefix. For teacher evidence $R$, define
\begin{equation}
\begin{aligned}
q^R_{t,i}&=q_{\mathrm T}(y_{t,i}\mid h_t,R,y_{t,<i}),\\
A^R_{t,i}&=\operatorname{sg}\!\left[
\log q^R_{t,i}-\log p_\theta(y_{t,i}\mid h_t,y_{t,<i})\right].
\end{aligned}
\label{eq:conditioned_signal}
\end{equation}
Training uses $R=R_t$; $\operatorname{sg}$ stops gradients through the
signal, whose value uses the current student forward pass.
We optimize a clipped policy-gradient surrogate \citep{schulman2017ppo},
with $\rho_{t,i}=p_\theta(y_{t,i}\mid h_t,y_{t,<i})/
p_{\mathrm{old}}(y_{t,i}\mid h_t,y_{t,<i})$:
\begin{equation}
\mathcal L=-\sum_{t,i}w_{t,i}\min\!\left\{
\rho_{t,i}A^{R_t}_{t,i},
\operatorname{clip}(\rho_{t,i},1-\epsilon,1+\epsilon)A^{R_t}_{t,i}\right\}.
\label{eq:opd}
\end{equation}
Here $\epsilon$ is the clipping radius and $w_{t,i}$ implements response
masking and episode normalization (Appendix~\ref{app:training_configuration});
batch and episode indices are suppressed.
The rollout policy $p_{\mathrm{old}}$ enters the ratio, whereas
Eq.~\ref{eq:conditioned_signal} uses the current actor. The batch closes with
\begin{equation}
\theta\leftarrow\operatorname{Update}(\theta;\mathcal L_{\mathcal B_\tau}),
\label{eq:student_update}
\end{equation}
where $\mathcal L_{\mathcal B_\tau}$ aggregates Eq.~\ref{eq:opd} over the
collected episodes. Only the student is updated; the teacher and offline
library remain fixed. Algorithm~\ref{alg:gcopd} in
Appendix~\ref{app:retrieval_algorithm} gives the complete control flow;
Appendices~\ref{app:evidence_shift} and~\ref{app:gradient} analyze the supervision signal.
Deployment requires neither teacher calls nor graph retrieval.

\section{Experiments}
\label{sec:results}
\paragraph{Case study: execution context changes feedback.}
\label{sec:action_case}
\label{ex:scoring_case}
\begingroup
\definecolor{caseaccent}{HTML}{000000}
\definecolor{casemuted}{HTML}{66717B}
\noindent In ScienceWorld, the student must find a living thing, focus on it, and move it to the blue box in the living room. The complete seven-action episode below starts in the art studio; step 4 is the scored decision.

\smallskip
\noindent\begin{minipage}{\linewidth}
\small
\setlength{\tabcolsep}{4pt}
\renewcommand{\arraystretch}{1.12}
\begin{tabular}{@{}p{.64in}p{4.70in}@{}}
\toprule
Student & \texttt{go hallway} $\rightarrow$ \texttt{go living room} $\rightarrow$ \texttt{look around} (sees a desk with a drawer).\\
Step 4 & \textcolor{caseaccent}{\texttt{look in drawer}}: ``The drawer isn't open, so you can't see inside.''\\
Steps 5--7 & \texttt{open drawer} (opens) $\rightarrow$ \texttt{look in drawer} (empty) $\rightarrow$ \texttt{focus on book}. \textbf{Failure.}\\
\midrule
Source & Same three-action prefix; the matched entry is \texttt{open drawer}, preceded by a rejected collection attempt to \texttt{look in drawer}. It opens and inspects the drawer (empty), examines the book and drawing, then visits the greenhouse, focuses on and picks up a pea plant, and carries it to the blue box. \textbf{Success.}\\
\bottomrule
\end{tabular}
\end{minipage}

\smallskip
\noindent\textbf{Scoring the original step-4 output:} \texttt{\textcolor{caseaccent}{look} in drawer}.
\begin{center}
\small
\setlength{\tabcolsep}{12pt}
\begin{tabular}{@{}lccc@{}}
\toprule
Teacher context & \texttt{look} & \texttt{in} & \texttt{drawer}\\
\midrule
Vanilla OPD & 97.69\% & \textcolor{casemuted}{99.96\%} & \textcolor{casemuted}{99.95\%}\\
GC-OPD & \textcolor{caseaccent}{26.81\%} & \textcolor{casemuted}{99.77\%} & \textcolor{casemuted}{99.99\%}\\
\bottomrule
\end{tabular}
\end{center}
\noindent Execution context reduces teacher support for \texttt{look} by \textbf{70.88 percentage points} at the rejected inspection step. Both conditions score identical student tokens with the original thought and response prefix. GC-OPD receives the complete student hindsight and source record; the display condenses observations and the source continuation. The probability change reflects their combined effect: the student hindsight itself includes the closed-drawer feedback.
\endgroup

\subsection{Experimental Setup}
We evaluate household interaction in ALFWorld, multistage scientific tasks
in ScienceWorld, and product search in WebShop
\citep{shridhar2020alfworld,wang2022scienceworld,yao2022webshop}.
Table~\ref{tab:main} compares vanilla OPD, TCOD-B2F, TCOD-F2B, FTB, and TurnOPD
under original and task-trained teachers where available.
Each trained model is evaluated with four inference seeds at temperature
0.4 and top-$p=1$. We report success, environment progress or reward, and
mean decisions over all tasks; $\pm$ denotes the sample SD across inference
seeds.

\textbf{Students share the acting protocol within each environment.}
Public prompts contain the task, observation, and most recent 5/5/2
observation--action pairs for ScienceWorld/ALFWorld/WebShop, with decision
limits of 30/30/15. Rejected or malformed actions consume a decision.
ScienceWorld has 1,661 evaluation tasks and reports best-progress Score;
ALFWorld has 140 Seen and 134 Unseen tasks, with rounds pooled over both;
WebShop has 500 tasks and reports final reward $\times100$.
Interaction and evaluation details appear in Appendix~\ref{app:evaluation_settings}.

\paragraph{Training schedule.}
For ScienceWorld, vanilla OPD has a two-epoch budget; GC-OPD uses one
vanilla OPD epoch followed by one GC epoch. Teachers remain frozen.
Figure~\ref{fig:homepage_teaser}(a) shows validation curves;
Appendix~\ref{app:training_configuration} gives training parameters.

\subsection{Main Results}
\begingroup
\setlength{\intextsep}{4pt}
\setlength{\abovecaptionskip}{3pt}
\begin{table}[!htbp]
\caption{\textbf{Main results across ScienceWorld, ALFWorld, and WebShop.}
(a) Two ScienceWorld student sizes. (b) Two ALFWorld teacher conditions and WebShop. Values are mean$\pm$SD over four inference seeds; bold/underline mark the best/second-best trained students per environment, size, and metric across teacher conditions. $^*$ denotes planner/oracle records used during training. The ALFWorld untrained student is repeated for comparison. Dashes denote unavailable or inapplicable entries.}
\label{tab:main}\label{tab:alfworld}\label{tab:webshop}
\begingroup
\fontsize{8}{9.5}\selectfont
\renewcommand{\meanstd}[2]{\mbox{\makebox[20pt][r]{{\fontsize{7.5}{8.5}\selectfont#1}}\makebox[14pt][l]{{\fontsize{5}{5.5}\selectfont\color{black!65}$\pm$#2}}}}
\newcommand{\meanonly}[1]{\mbox{\makebox[20pt][r]{{\fontsize{7.5}{8.5}\selectfont#1}}\makebox[14pt][l]{}}}
\newcommand{\stathead}[1]{\makebox[20pt][c]{\shortstack{#1}}\makebox[14pt]{}}
\newcommand{\modelpair}[1]{{\fontsize{6.5}{7.5}\selectfont\mbox{#1}}}
\noindent\textbf{(a) ScienceWorld}\par\vspace{2pt}
\begingroup
\centering
\setlength{\tabcolsep}{3pt}
\begin{tabularx}{\linewidth}{@{}Xccc@{}}
\toprule
Teacher policy: Qwen3-32B & \stathead{SR $\uparrow$} & \stathead{Score $\uparrow$} & \stathead{Rounds $\downarrow$}\\
\midrule
Original & \meanstd{29.74}{0.30} & \meanstd{57.24}{0.27} & \meanstd{17.87}{0.03}\\
GRPO & \meanstd{48.43}{0.56} & \meanstd{70.57}{0.50} & \meanstd{15.74}{0.13}\\
\bottomrule
\end{tabularx}\par
\vspace{0.4em}
\setlength{\tabcolsep}{1.8pt}
\begin{tabularx}{\linewidth}{@{}Xcccccc@{}}
\toprule
&\multicolumn{3}{c}{Student: Qwen3-1.7B}&\multicolumn{3}{c}{Student: Qwen3-4B}\\
\cmidrule(lr){2-4}\cmidrule(lr){5-7}
Method & \stathead{SR $\uparrow$} & \stathead{Score $\uparrow$} & \stathead{Rounds $\downarrow$}
& \stathead{SR $\uparrow$} & \stathead{Score $\uparrow$} & \stathead{Rounds $\downarrow$}\\
\midrule
Untrained student & \meanstd{3.09}{0.39} & \meanstd{9.31}{0.15} & \meanstd{22.55}{0.22} & \meanstd{15.01}{0.30} & \meanstd{35.46}{0.36} & \meanstd{17.27}{0.13}\\
\midrule\multicolumn{7}{@{}l}{\textit{Scoring teacher: Original Qwen3-32B}}\\
Vanilla OPD & \meanstd{22.31}{0.40} & \meanstd{52.79}{0.35} & \meanstd{19.66}{0.15} & \meanstd{24.70}{0.35} & \meanstd{53.47}{0.52} & \meanstd{17.24}{0.15}\\
TCOD-B2F\textsuperscript{*} & \meanstd{21.06}{0.94} & \meanstd{49.20}{0.53} & \meanstd{17.76}{0.15} & \meanstd{23.13}{0.33} & \meanstd{52.30}{0.04} & \meanstd{18.77}{0.15}\\
TCOD-F2B & \meanstd{23.34}{0.51} & \meanstd{50.20}{0.46} & \meanstd{16.91}{0.11} & \meanstd{23.60}{0.80} & \meanstd{53.05}{0.64} & \meanstd{18.78}{0.20}\\
FTB\textsuperscript{*} & \meanstd{22.70}{0.60} & \meanstd{51.35}{0.40} & \meanstd{18.11}{0.12} & \meanstd{23.21}{0.25} & \meanstd{52.60}{0.23} & \meanstd{18.39}{0.18}\\
TurnOPD & \meanstd{23.84}{0.41} & \meanstd{52.63}{0.62} & \meanstd{18.51}{0.13} & \meanstd{23.92}{0.83} & \meanstd{54.32}{0.70} & \meanstd{17.72}{0.10}\\
\method & \meanstd{\underline{46.18}}{1.13} & \meanstd{\underline{68.10}}{0.59} & \meanstd{\textbf{13.04}}{0.10} & \meanstd{\underline{48.78}}{0.58} & \meanstd{68.09}{0.19} & \meanstd{\underline{11.27}}{0.16}\\
\method{} + GA\textsuperscript{*} & \meanstd{\textbf{53.61}}{0.48} & \meanstd{\textbf{73.55}}{0.48} & \meanstd{\underline{13.07}}{0.18} & \meanstd{\textbf{54.68}}{0.62} & \meanstd{\textbf{71.90}}{0.44} & \meanstd{\textbf{11.06}}{0.12}\\
\midrule\multicolumn{7}{@{}l}{\textit{Scoring teacher: GRPO Qwen3-32B}}\\
Vanilla OPD & \meanstd{38.70}{0.46} & \meanstd{63.36}{0.25} & \meanstd{16.29}{0.05} & \meanstd{46.66}{0.36} & \meanstd{\underline{71.44}}{0.58} & \meanstd{14.64}{0.09}\\
TCOD-B2F\textsuperscript{*} & \meanstd{29.91}{0.97} & \meanstd{57.55}{0.79} & \meanstd{16.90}{0.08} & \meanstd{37.84}{0.52} & \meanstd{63.32}{0.08} & \meanstd{16.38}{0.08}\\
TCOD-F2B & \meanstd{27.75}{0.27} & \meanstd{56.72}{0.20} & \meanstd{17.20}{0.05} & \meanstd{40.13}{0.17} & \meanstd{64.88}{0.53} & \meanstd{16.21}{0.15}\\
FTB\textsuperscript{*} & \meanstd{31.58}{0.71} & \meanstd{58.38}{0.58} & \meanstd{16.67}{0.13} & \meanstd{36.56}{0.29} & \meanstd{62.42}{0.57} & \meanstd{16.70}{0.10}\\
TurnOPD & \meanstd{35.75}{0.78} & \meanstd{60.91}{0.39} & \meanstd{16.32}{0.09} & \meanstd{44.18}{0.96} & \meanstd{68.62}{0.84} & \meanstd{16.05}{0.08}\\
\bottomrule
\end{tabularx}\par\endgroup
\par\vspace{4pt}\noindent\textbf{(b) ALFWorld and WebShop}\par\vspace{2pt}
\begingroup

\setlength{\tabcolsep}{0.7pt}
\renewcommand{\arraystretch}{1.02}

\newcolumntype{N}{>{\raggedleft\arraybackslash}p{34pt}}
\begin{tabularx}{\linewidth}{@{}X*{9}{N}@{}}
\toprule
&\multicolumn{3}{c}{\textbf{ALFWorld}}&\multicolumn{3}{c}{\textbf{ALFWorld}}&\multicolumn{3}{c}{\textbf{WebShop}}\\
&\multicolumn{3}{c}{\modelpair{Qwen3-32B $\to$ Qwen3-1.7B}}&\multicolumn{3}{c}{\modelpair{Qwen3-32B GRPO $\to$ Qwen3-1.7B}}&\multicolumn{3}{c}{\modelpair{Qwen3.8-27B $\to$ Qwen3.5-0.8B}}\\
\cmidrule(lr){2-4}\cmidrule(lr){5-7}\cmidrule(l){8-10}
Method & \stathead{Seen\\SR $\uparrow$} & \stathead{Unseen\\SR $\uparrow$} & \stathead{Rounds\\$\downarrow$} & \stathead{Seen\\SR $\uparrow$} & \stathead{Unseen\\SR $\uparrow$} & \stathead{Rounds\\$\downarrow$} & \stathead{SR\\$\uparrow$} & \stathead{Score\\$\uparrow$} & \stathead{Rounds\\$\downarrow$}\\
\midrule
Teacher & \meanstd{46.96}{1.58} & \meanstd{49.81}{2.23} & \meanstd{21.64}{0.40} & \meanstd{83.57}{2.10} & \meanstd{80.22}{1.29} & \meanstd{14.05}{0.23} & \meanstd{29.50}{0.82} & \meanstd{57.54}{0.47} & \meanstd{6.34}{0.12}\\
\addlinespace[1.5pt]
Untrained student & \meanstd{7.86}{1.75} & \meanstd{9.14}{1.54} & \meanstd{28.35}{0.25} & \meanstd{7.86}{1.75} & \meanstd{9.14}{1.54} & \meanstd{28.35}{0.25} & \meanstd{2.45}{0.82} & \meanstd{5.18}{0.77} & \meanstd{14.51}{0.05}\\
\addlinespace[1.5pt]
Vanilla OPD & \meanstd{44.29}{2.47} & \meanstd{53.36}{2.55} & \meanstd{21.69}{0.18} & \meanstd{82.86}{2.67} & \meanstd{79.48}{2.24} & \meanstd{13.71}{0.31} & \meanstd{29.10}{0.84} & \meanstd{55.61}{1.18} & \meanstd{6.82}{0.08}\\
TCOD-B2F\textsuperscript{*} & \meanstd{46.25}{2.94} & \meanstd{55.60}{2.62} & \meanstd{21.51}{0.38} & \meanstd{87.50}{0.41} & \meanstd{85.07}{1.49} & \meanstd{12.48}{0.12} & \meanstd{31.30}{1.55} & \meanstd{56.01}{1.29} & \meanstd{6.86}{0.13}\\
TCOD-F2B & \meanstd{43.57}{2.26} & \meanstd{51.87}{3.93} & \meanstd{22.10}{0.32} & \meanstd{85.71}{1.93} & \meanstd{83.21}{1.55} & \meanstd{12.69}{0.11} & \meanstd{30.25}{1.47} & \meanstd{56.71}{1.28} & \meanstd{6.64}{0.11}\\
FTB\textsuperscript{*} & \meanstd{48.39}{3.89} & \meanstd{53.92}{2.14} & \meanstd{21.02}{0.16} & \meanstd{86.07}{2.22} & \meanstd{81.90}{1.54} & \meanstd{13.23}{0.18} & \meanstd{30.90}{0.48} & \meanstd{57.63}{0.63} & \meanstd{\underline{6.54}}{0.11}\\
TurnOPD & \meanstd{38.57}{2.54} & \meanstd{47.01}{1.61} & \meanstd{22.92}{0.36} & \meanstd{80.36}{1.37} & \meanstd{76.12}{2.44} & \meanstd{14.35}{0.39} & \meanstd{27.50}{0.66} & \meanstd{54.29}{0.52} & \meanstd{6.77}{0.11}\\
\method & \meanstd{\underline{91.61}}{0.68} & \meanstd{\underline{85.26}}{0.71} & \meanstd{\underline{10.93}}{0.19} & \meanonly{--} & \meanonly{--} & \meanonly{--} & \meanstd{\underline{37.65}}{0.93} & \meanstd{\underline{60.14}}{1.06} & \meanstd{6.66}{0.04}\\
\method{} + GA\textsuperscript{*} & \meanstd{\textbf{96.07}}{0.92} & \meanstd{\textbf{93.47}}{0.71} & \meanstd{\textbf{9.60}}{0.11} & \meanonly{--} & \meanonly{--} & \meanonly{--} & \meanstd{\textbf{39.90}}{1.06} & \meanstd{\textbf{62.96}}{1.08} & \meanstd{\textbf{6.37}}{0.07}\\
\bottomrule\end{tabularx}\par\endgroup

\endgroup
\end{table}

\endgroup
\paragraph{Higher success than distillation from GRPO-trained teachers.}
With the original teacher, GC-OPD achieves higher mean success than every
evaluated GRPO-teacher OPD baseline on ScienceWorld and ALFWorld at matched student sizes
(Table~\ref{tab:main}). On ScienceWorld, the 1.7B and 4B students reach
46.18\% and 48.78\%, exceeding the strongest GRPO-teacher baselines,
both vanilla OPD, at 38.70\% and 46.66\%. On ALFWorld, GC-OPD reaches
91.61\% Seen and 85.26\% Unseen success, versus 87.50\% and 85.07\%
for the strongest GRPO-teacher baseline, TCOD-B2F. The Unseen advantage
is 0.19 percentage points in the reported mean. These comparisons support
improving the teacher's scoring context as an alternative to optimizing
its parameters. GC-OPD also improves WebShop success
from 29.10\% to 37.65\% over vanilla OPD.

\paragraph{Higher success with fewer interactions.}
Against the best GRPO-teacher baselines above, the ScienceWorld 1.7B
student uses 13.04 mean rounds versus 16.29; the 4B student uses 11.27
versus 14.64. ALFWorld rounds fall from 12.48 to 10.93. GC-OPD thus
achieves higher success with fewer decisions per task on average.
Mean rounds include both successful and failed evaluation episodes.

\subsection{Ablation and Analysis}
\label{sec:ablation_analysis}
\label{sec:reference_ablation}
\paragraph{Student hindsight and reference selection.}
The hindsight-only control uses the same vanilla OPD parent and continuation
budget as GC-OPD, providing the complete student execution and outcome without external
records (Appendix~\ref{app:training_configuration}). It reaches 31.64\% success, compared with 22.31\% for vanilla
OPD and 46.18\% for GC-OPD (Table~\ref{tab:reference_alignment}(a)).
Student hindsight alone therefore falls 14.54 points below the full method.
To examine external-reference selection, the minimum-prefix control uses
the same K16 pool and student hindsight, but fixes successful and failed
records with the shortest shared action prefix throughout each episode.
Its 33.70\% success leaves a 12.48-point gap to GC-OPD. These results show
that the graph helps select useful execution records for supervising
student decisions.

In the full GC stage (3,017 episodes, 42,537 decisions), current locators
match indexed visits at 37.25\% of decisions. After ranking and retention,
16.23\% receive a current-state successful reference and 25.82\% receive
one through an earlier student-state anchor. Historical anchors thus
provide more retained successful references than current-state lookup,
extending reference availability beyond direct success support.

\paragraph{Outcome composition.}
We test whether failed external records add value beyond successful
references and student hindsight. From the same vanilla OPD parent,
training on all 3,017 tasks with the K16 graph but omitting external
failed references yields 43.68\% success and 13.27 mean rounds, versus
46.18\% and 13.04 with both source outcomes
(Table~\ref{tab:reference_alignment}(a)). Removing failures loses 2.50
percentage points despite preserving successful sources and student
hindsight: useful external evidence extends beyond successful solutions.

Across the default run's 42,537 decisions, external references contain
successful records only (17.68\%), both outcomes (26.00\%), failed records
only (47.34\%), or no record (8.98\%), including labelled fallbacks.
Every category retains the complete student execution. Failed records
provide both contrasts to successful solutions and the only external
action-consequence evidence for nearly half of these decisions.

\begin{table}[!htbp]
\caption{\textbf{Execution-evidence ablations.} ScienceWorld, 1.7B students, no planner sources; mean$\pm$SD across four inference seeds. (a) Original 32B scorer. Hindsight-only supplies the student's complete execution without external records; the reference-based variants use $K=16$ and retain student hindsight. ``Success only'' removes external failed records. (b) Student SR by generator/scorer; coverage is training-task trusted-success coverage. The 8B-scorer source exchange is matched; the 32B row reports selected pipelines.}
\label{tab:reference_alignment}\label{tab:teacher_exchange_main}
\centering\small
\begin{minipage}[t]{.56\linewidth}
\centering\textbf{(a) Teacher context}\par\smallskip
\setlength{\tabcolsep}{3pt}
\begin{tabular}{@{}lcc@{}}
\toprule
Configuration & SR & Rounds\\
\midrule
Vanilla OPD & \meanstd{22.31}{0.40} & \meanstd{19.66}{0.15}\\
Hindsight-only & \meanstd{31.64}{0.57} & \meanstd{13.83}{0.13}\\
Minimum-prefix & \meanstd{33.70}{0.63} & \meanstd{16.76}{0.10}\\
GC: success only & \meanstd{43.68}{0.56} & \meanstd{13.27}{0.18}\\
GC: success + failure & \meanstd{46.18}{1.13} & \meanstd{13.04}{0.10}\\
\bottomrule
\end{tabular}
\end{minipage}\hfill
\begin{minipage}[t]{.42\linewidth}
\centering\textbf{(b) Generator / scorer}\par\smallskip
\setlength{\tabcolsep}{3pt}
\begin{tabular}{@{}lcc@{}}
\toprule
Scorer & 8B source & 32B source\\
\midrule
8B & \meanstd{27.03}{0.44} & \meanstd{20.97}{0.34}\\
32B & \meanstd{42.49}{1.11} & \meanstd{46.18}{1.13}\\
\midrule
Coverage & 48.79\% & 59.07\%\\
\bottomrule
\end{tabular}
\end{minipage}
\end{table}

\begin{figure}[!htbp]
\centering
\includegraphics[width=.82\linewidth]{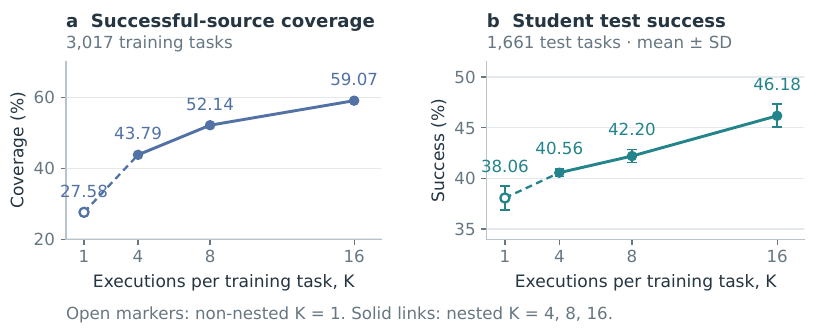}
\caption{\textbf{Source budget and student performance.} ScienceWorld, 1.7B student and original 32B teacher. (a) Trusted-success coverage on 3,017 training tasks. (b) Test SR on 1,661 tasks (mean$\pm$SD, four inference seeds). Axes differ; $K=1$ is separately preselected, while $K=4,8,16$ are nested.}
\label{fig:evidence_results}
\end{figure}

\paragraph{Source budget and augmentation.}
With the original 32B generator and scorer fixed, increasing $K$ from 1 to 16
raises successful-source coverage from 27.58\% to 59.07\% and student
test success from 38.06\% to 46.18\%
(Figure~\ref{fig:evidence_results}). The 8.12-point student gain shows
that repeated sampling supplies useful training evidence without changing
the teacher's parameters. Adding planner executions through the same
retrieval interface yields 54.68\% success for the ScienceWorld 4B
student, 93.47\% on ALFWorld Unseen, and 39.90\% on WebShop
(Table~\ref{tab:main}). The same retrieval interface can thus use
complementary execution sources, extending the benefit beyond repeated
attempts by one teacher (Appendix~\ref{app:source_collection}).

\paragraph{Source generator and scoring teacher.}
With the 8B scorer fixed, replacing 8B-generated records with 32B records
raises trusted-success coverage from 48.79\% to 59.07\%, yet lowers
student success from 27.03\% to 20.97\%
(Table~\ref{tab:teacher_exchange_main}(b)). The branches share initialization,
continuation budget, and selection rule. In this comparison, higher
successful-source coverage does not translate into better student performance,
suggesting that coverage alone does not fully capture the value of execution
records for distillation.

\subsection{Computational Cost}
\label{sec:compute}
Figure~\ref{fig:homepage_teaser}(b) compares the ScienceWorld 1.7B routes:
K16 collection plus GC-OPD (46.18\% success), and GRPO teacher training
plus vanilla OPD (38.70\%). On 64 H20 GPUs, collection takes 3.32 hours
and teacher training 18.27 hours. Two-epoch student training on eight
H20 GPUs takes 8.60 and 5.91 hours, respectively. The measured stages
sum to 281.27 versus 1,216.59 GPU-hours
(Appendix~\ref{app:compute_details}); lower preparation cost offsets GC-OPD's
higher student-training cost. GC-OPD also avoids teacher-side gradient
and optimizer storage.

\section{Conclusion}
At matched student sizes, GC-OPD with off-the-shelf teachers achieves higher
mean success than every evaluated GRPO-teacher OPD baseline on ScienceWorld
and ALFWorld, while also improving WebShop performance over vanilla OPD.
The execution graph links trajectories through shared states and retrieves
useful references for student supervision. Complete source histories and
student hindsight condition feedback on original responses, supporting
graph-guided distillation without task-specific teacher optimization.
\label{mainend}
\bibliography{references}
\bibliographystyle{iclr2027_conference}
\clearpage
\appendix
\section{Training and evaluation settings}
\label{app:protocol_adaptations}
\label{app:protocol}

\subsection{Interaction, metrics, and data partitions}
\label{app:evaluation_settings}
\paragraph{Dataset partitions.}
ALFWorld uses a 3,553-task training pool and the official 140 Seen and
134 Unseen evaluation tasks. ScienceWorld uses 3,017 training, 305 development, and
1,661 test instances. Its development tasks are held out from the official
training split; its test tasks are drawn from the official development split.
WebShop uses 3,000 unique training goals, 200
development goals, and the official 500 test goals. Execution-source
libraries are constructed from training tasks.

\paragraph{Agent inputs and evaluation.}
Each decision receives a newly constructed user message containing the task,
the current observation, and a bounded history of observation--action pairs.
Each historical observation precedes its paired action. The response to that
action becomes the next current observation. Earlier assistant responses,
historical thoughts, and an additional student memory module are not appended.
ScienceWorld additionally supplies possible action templates and object
names from the environment interface. ALFWorld supplies admissible
commands, and WebShop supplies the current search/click actions.
The compared students receive the same environment-provided information.
The same input construction and action parser are used during student
rollouts and evaluation within each environment.

\begin{table}[!htbp]
\centering\small
\caption{Environment-specific interaction settings. Token limits are per
decision; teacher-reference limits apply to graph-conditioned scoring, not
to the deployed student. Rounds denotes the mean number of agent decisions
over all evaluated tasks.}
\label{tab:environment_protocols}
\begin{tabularx}{\linewidth}{@{}Xccc@{}}
\toprule
Setting & ScienceWorld & ALFWorld & WebShop\\
\midrule
Evaluation tasks & 1,661 & 140 Seen / 134 Unseen & 500\\
History pairs & 5 & 5 & 2\\
Maximum decisions & 30 & 30 & 15\\
Student prompt tokens & 10,240 & 10,240 & 63,488\\
Student response tokens & 512 & 512 & 2,048\\
Graph-teacher context tokens & 40,513 & 40,513 & 262,144\\
Evaluation temperature & 0.4 & 0.4 & 0.4\\
Evaluation top-$p$ & 1 & 1 & 1\\
\bottomrule
\end{tabularx}
\end{table}

Task success is the environment's completed-success indicator. ScienceWorld
Score is the highest progress score reached in an episode, following
TCOD's released implementation \citep{wang2026tcod}.
WebShop reports final reward multiplied by 100, with success requiring a
completed episode and reward at least $1-10^{-9}$. ALFWorld's recorded score
is binary success, so it does not provide a separate continuous-score column.
Rounds counts model decisions, including malformed responses and rejected
actions, rather than simulator ticks or only successful executions.
Reported means and sample standard deviations use four inference seeds for
one fixed trained model, with standard-deviation denominator $4-1$.

\paragraph{Multiple-attempt teacher reference.}
Table~\ref{tab:teacher_pass16} reports the original teachers' recorded
sixteen-attempt evaluation banks, separate from the four-seed
single-episode evaluations in Table~\ref{tab:main}. Pass@16 is the
fraction of tasks with at least one successful attempt; Score@16 averages
the highest episode Score per task. This is a sampling reference, not an
upper bound on the trained student.
\begin{table}[!htbp]
\centering\small
\caption{Original-teacher performance with sixteen attempts per task.
ALFWorld has binary scores, so a separate Score@16 is omitted.}
\label{tab:teacher_pass16}
\begin{tabularx}{\linewidth}{@{}Xrrr@{}}
\toprule
Environment & Tasks & Pass@16 (\%) & Score@16\\
\midrule
ScienceWorld & 1,661 & 66.83 & 83.11\\
ALFWorld Seen & 140 & 77.86 & --\\
ALFWorld Unseen & 134 & 82.09 & --\\
WebShop & 500 & 50.60 & 75.60\\
\bottomrule
\end{tabularx}
\end{table}

\paragraph{Prompt and parser adaptations.}
Our task templates build on TCOD, while the bounded observation--action
history settings follow the interaction design used with OPID
\citep{wang2026tcod,yang2026opid}. We use \texttt{<thought>} in place of
\texttt{<think>}, disable the tokenizer's native thinking mode, and retain
any generated custom thought as part of the current response. Parsing
accepts an optional closed thought followed by exactly one closed
\texttt{<action>\ldots</action>} block. Multiple action blocks, unclosed
tags, and non-whitespace text outside the permitted blocks are invalid.
ScienceWorld additionally allows an empty action when its ambiguity prompt
explicitly requests a blank cancellation; ALFWorld and WebShop reject
empty actions.

A malformed response does not trigger an environment action. It consumes
one decision and returns explicit format-error feedback, after which the
agent may continue within its remaining budget. A well-formed action that
the environment rejects also consumes a decision and retains the actual
feedback. No fallback action is extracted from the response's final
characters, and evaluation grants no free format retries. Thus our
reproductions use a shared adapted interaction protocol; method-specific
teacher supervision, curriculum, and bridge mechanisms remain separate
from the student's history representation.

\paragraph{WebShop model identities.}
The student is \href{https://huggingface.co/Qwen/Qwen3.5-0.8B/tree/2fc06364715b967f1860aea9cf38778875588b17}{Qwen3.5-0.8B} at revision
\texttt{2fc06364715b}; the teacher is
\href{https://huggingface.co/Qwen/Qwen3.8-27B/tree/1d4bf0f2ff6012fd82039f2fa52739d0dd7c60c0}{Qwen3.8-27B} at revision
\texttt{1d4bf0f2ff60}. Both use the compatible \texttt{qwen3\_5} architecture
configuration; the model names and revisions are verified against the
archived download metadata. 

\subsection{Training configuration}
\label{app:training_configuration}
Table~\ref{tab:common_gc_configuration} summarizes the shared settings for
primary GC-OPD training.

\begin{table}[!htbp]
\centering\small
\caption{Common training configuration.}
\label{tab:common_gc_configuration}
\begin{tabularx}{\linewidth}{@{}XlXl@{}}
\toprule
Parameter & Value & Parameter & Value\\
\midrule
Optimizer & AdamW & Learning rate & $10^{-5}$\\
Adam betas & $(0.9,0.999)$ & Weight decay & 0\\
LR schedule & Constant & LR warmup steps & 0\\
Task batch size & 32 & Rollouts per task & 1\\
Training temperature & 1.0 & Top-$p$ & 1.0\\
Gradient clipping & 1.0 & OPD clipping & 0.2\\
Actor passes per batch & 1 & Teacher parameters & Frozen\\
\bottomrule
\end{tabularx}
\end{table}

The loss averages over response tokens within each episode and then equally
over episodes: $w_{e,t,i}=1/(N L_e)$, where $N$ is the number of episodes
and $L_e$ is episode $e$'s total response-token count. The teacher--student
log-probability feedback has coefficient one; task-reward advantages are disabled.

For TurnOPD, we follow the original paper's adaptive rollout-depth and
progressive turn-normalization design \citep{zhou2026turnopd}.

\paragraph{Hindsight-only control.}
This control resumes the same vanilla OPD parent as ScienceWorld 1.7B
GC-OPD and uses the same continuation budget, retaining student hindsight
and disabling all external records. All four inference seeds evaluate
the same fixed model in each condition.

\section{Method details}
\label{app:method_details}
\label{app:execution_evidence}
This appendix details how recorded executions are organized, selected as
teacher context, and used to form the distillation signal.
\subsection{Execution sources}
\label{app:source_collection}
\paragraph{Teacher executions.}
Source libraries use training tasks and remain fixed during the GC stage.
In ScienceWorld, the original teacher records up to 30 accepted actions
per execution, with temperature 0.4, top-$p$ 1, top-$k$ 20, a 10,240-token
prompt limit, and a 512-token response limit. Each action position permits
at most five sampling attempts: format errors are resampled without an
environment action, while environment rejection triggers reset and verified
replay of the accepted prefix before resampling. The collection history
accumulates accepted user messages and action-only replies. Student
rollouts and evaluation instead count every decision without free retries.

ALFWorld collection uses the H5 single-message protocol, 30 decisions,
temperature 0.4, top-$p$ 1, top-$k$ $-1$, and 512 response tokens; malformed
decisions consume a turn. WebShop uses H2 and 2,048 response tokens and
permits same-state format resampling during collection. Records retain
their actions, observations, feedback, outcomes, and source identities.
Source-library success coverage is reported separately from single-execution
teacher evaluation; reference selection and rendering follow
Appendix~\ref{app:retrieval_algorithm}.

\paragraph{Planner and oracle executions.}
Graph augmentation adds executed planner or oracle records, with their
origins and observed outcomes, to the teacher library before indexing.
ScienceWorld executes its built-in gold-path planner with a 200-action
collection limit, so a source may exceed the student's 30-decision
horizon. ALFWorld admits replay-verified successful TextWorld plans within
the interaction budget \citep{cote2018textworld}; tasks without an admitted
plan remain in training. WebShop uses a rule-based oracle with training-task
product and attribute information. In our baseline reproductions,
TCOD-B2F and FTB use planner prefixes; FTB also uses teacher bridges with
future-continuation verification. TCOD-F2B and TurnOPD do not use planner
prefixes.

The ScienceWorld 1.7B augmented variant additionally labels the final
response when the episode ends naturally with a finite negative score,
excluding horizon and technical stops. The ScienceWorld 4B and ALFWorld
variants do not add this note.

\subsection{State descriptors and matching}
\label{app:state_descriptors}

ScienceWorld combines a canonical physical-configuration hash with the
feedback-derived focus name, ordered goal flags, and parser mode/options.
Continuous-valued quantities such as temperature are discretized for
matching. ALFWorld hashes the task-file identifier, sorted grounded PDDL
facts, and won/lost flags. WebShop hashes page type, ordered search keywords,
result-page index, product identifier (ASIN), sorted selected options,
and item subpage; terminal visits use the done marker, ASIN, and selected
options. WebShop reward and previously visited products are retained as
metadata and do not enter the locator. These task-specific descriptors
support retrieval and are not student inputs.

Every visit retains its source identity and committed-step position,
including actions that leave the locator unchanged. In the ScienceWorld
catalog, key-changing transitions are stored as edges and unchanged-key
events as node records; both contribute temporal transitions to $H_x$
and its projection in Eq.~\ref{eq:graph_nodes}. Unreliable captures cannot
match. Noncommitted collection retries remain labelled metadata.

\subsection{Reference retrieval and context construction}
\label{app:retrieval_algorithm}
\label{app:retrieval_details}

\newcounter{gcopdalgorithm}
\begin{figure}[!htbp]
\refstepcounter{gcopdalgorithm}
\label{alg:gcopd}
\hrule\vspace{4pt}
\noindent\textbf{Algorithm \thegcopdalgorithm: Graph-conditioned on-policy distillation}
\vspace{3pt}\hrule\vspace{4pt}
\small
\noindent\textbf{Inputs:} $\mathcal X,\ p_\theta,\ q_{\mathrm{gen}},\ q_{\mathrm T},\ \mathrm{env}$.
\par\vspace{3pt}
\renewcommand{\arraystretch}{1.14}
\begin{tabularx}{\linewidth}{@{}>{\scriptsize\color{gray}}r@{\hspace{0.65em}}X@{}}
1 & $\{\mathcal D_x,H_x,G_x\}_{x\in\mathcal X}\leftarrow\operatorname{CollectIndex}(\mathcal X,q_{\mathrm{gen}})$\\
2 & $\theta\leftarrow\operatorname{OPDWarmStart}(\theta,q_{\mathrm T})$\\
3 & \textbf{for} GC-stage training batch $B\subseteq\mathcal X$:\\
4 & \hspace*{1em}$p_{\mathrm{old}}\leftarrow p_\theta$; $\mathcal B_\tau\leftarrow\operatorname{Rollout}(p_{\mathrm{old}},B)$\\
5 & \hspace*{1em}\textbf{for} $(\tau,t)\in\operatorname{Decisions}(\mathcal B_\tau)$:\\
6 & \hspace*{2em}$\mathcal A_t^+\leftarrow\operatorname{SuccessVisits}(z_t,\mathcal D_x)$ \hfill [Eq.~\ref{eq:recorded_support}]\\
7 & \hspace*{2em}\textbf{if} $\mathcal A_t^+\ne\varnothing$:\\
8 & \hspace*{3em}$\mathcal C_t^+\leftarrow\mathcal A_t^+\cup\{(\tau,t)\mid\operatorname{won}(\tau)=1\}$;
$v_t^*\leftarrow\arg\min_{v\in\mathcal C_t^+}d(v)$\\
9 & \hspace*{3em}$\mathcal S_t\leftarrow\operatorname{External}(v_t^*)$\\
10 & \hspace*{2em}\textbf{else}:\\
11 & \hspace*{3em}$\alpha_t\leftarrow\max\{u\le t:\mathcal A_u^+\ne\varnothing\}$; $\max\varnothing:=\bot$\\
12 & \hspace*{3em}$v_t^+\leftarrow$
$\arg\min_{v\in\mathcal A_{\alpha_t}^+}d(v)$ \textbf{if} $\alpha_t\ne\bot$
\textbf{else} $\operatorname{FallbackSuccess}(\mathcal D_x)$\\
13 & \hspace*{3em}$v_t^-\leftarrow\operatorname{FailedRef}(\tau,t,\mathcal D_x)$;
$\mathcal S_t\leftarrow\operatorname{Retain}_{\mathrm{env}}(v_t^+,v_t^-;\tau,t,G_x)$\\
14 & \hspace*{2em}$R_t\leftarrow\operatorname{Render}(\tau,t,\mathcal S_t)$\\
15 & \hspace*{1em}$\{A^{R_t}_{t,i}\}_{\mathcal B_\tau}\leftarrow\operatorname{Score}(q_{\mathrm T},p_\theta;\mathcal B_\tau,\{R_t\})$ \hfill [Eq.~\ref{eq:conditioned_signal}]\\
16 & \hspace*{1em}$\theta\leftarrow\operatorname{Update}(\theta;\mathcal L_{\mathcal B_\tau})$ \hfill [Eq.~\ref{eq:opd}]\\
\end{tabularx}
\par\vspace{3pt}
\footnotesize Helpers, retention gates, and empty-reference cases:
Appendix~\ref{app:retrieval_algorithm}.
\par\vspace{4pt}\hrule
\end{figure}

\paragraph{Candidate admission.}
Shared-node lookup returns $\mathcal I_x(z_t)$ in
Eq.~\ref{eq:node_visit_index}; eligibility tests form $\mathcal A_t^+$
as in Eq.~\ref{eq:recorded_support}.
A match requires a captured, matchable locator; in ScienceWorld it also
requires the same parser mode and option mapping. Matching alone does not
establish eligibility as a successful source. ScienceWorld and WebShop admit successful
suffixes only from records marked complete and successful, with no recorded
continuity gap and with consecutive feedback/observation agreement after
whitespace normalization. The entry decision must have an action or binding
and must not be marked rejected, format-invalid, rolled back, or ineligible;
this check applies to the entry, not every subsequent decision. Their terminal
successful visits admit empty suffixes. ALFWorld uses replay-checked catalogs
and requires a nonterminal entry with an executed, nonrejected, well-formed
action; terminal empty suffixes are not candidates.
An indexed current-state match means that a source visit matches before
the successful-suffix eligibility test. Such a match can therefore exist
even when $\mathcal A_t^+=\varnothing$; the retention rules below distinguish
these two conditions. These checks establish source-record integrity and
an environment/control association, not equality of acquired information.
Complete prefixes are retained so that differences in observations and
preparations remain visible to the scoring teacher.

\paragraph{Decision cost and ties.}
The cost $d(r,j)$ counts committed interaction decisions from source visit
$j$ to the recorded ending, including observation and binding decisions.
Noncommitted collection retries are excluded. Every recorded student turn
has unit cost, including malformed or rejected decisions. Successful-source
ties favor an external record over the student's continuation, then smaller
repetition index and earlier source position. Whole-source cost is $d(r,0)$.

\paragraph{Per-decision procedure.}
For each original response at decision $t$, the operators in
Algorithm~\ref{alg:gcopd} expand as follows; all searches stay within task $x$.
\begin{enumerate}
\setlength{\itemsep}{0pt}
\setlength{\parskip}{0pt}
\item If $\mathcal A_t^+\ne\varnothing$, rank its visits together with the
student's own continuation $(\tau,t)$ only if the episode succeeds.
For the winning visit $v$,
$\operatorname{External}(v)$ retains its complete external source or returns
no external record if the student's continuation wins the ranking.
\emph{This branch does not add a failed reference or invoke fallback retention.}
\item Only if $\mathcal A_t^+=\varnothing$, scan actual student visits from
$t$ back to $0$ for the latest successful-source anchor, with the same
suffix ranking. Without a shared success anchor,
$\operatorname{FallbackSuccess}(\mathcal D_x)$ selects an eligible complete
same-task success by whole-source cost and labels it unaligned; if none
exists, no trusted successful reference is returned. An absent anchor is
$\bot$ and is never used to index a candidate set or take an empty argmin.
Independently,
$\operatorname{FailedRef}$ finds the latest student position $u$ shared with
a failed record and minimizes $(|j-u|,d(r,0),\text{repetition},j)$ over its
source visits. Without a shared position, it minimizes whole-source cost
and repetition index and labels the record unaligned. Successful and failed
references may consequently use different historical anchors.
\item The fallback branch applies $\operatorname{Retain}_{\mathrm{env}}$.
A failed student retains available successful and failed references.
A successful student with no indexed current-state match uses itself alone.
If it has a current match but no eligible external success, available
historical or unaligned successful and failed references are retained
alongside the student's complete execution.
ScienceWorld and WebShop also permit readable raw records when trusted
references are unavailable, explicitly labelled unverified and unaligned;
these never enter successful-suffix ranking. No missing source is fabricated.
\item Render the complete student execution first, including feedback,
final outcome and the marked decision being scored. Follow it with at most
one successful and one failed complete source, preserving source identities,
verification, outcomes, and current/historical anchors or unaligned labels.
Selected successful records precede failed records in all environments. Preserve each record's own prefix,
suffix, actions, observations and feedback; historical thoughts are excluded.
Insert the evidence into the original user prompt and score the unchanged
response IDs with the fixed teacher.
\end{enumerate}

\paragraph{Rendered evidence.}
This excerpt preserves actual renderer labels; brackets denote omitted
variable fields in this illustration, not truncation during training.
The current action appears in hindsight; historical thoughts do not.
\begin{quote}\footnotesize\ttfamily
STUDENT'S COMPLETE ACTUAL ACTION/OBSERVATION EXECUTION\\
Student event [t] [CURRENT SCORED DECISION]\\
Recorded action: [original parsed student action]\\
Actual feedback: [recorded feedback]\\
{}[all earlier/later events and final outcome]\\
COMPLETE SOURCE ACTION/OBSERVATION EXECUTION: [source ID]\\
{}[verification, anchor, full prefix/suffix, outcome]
\end{quote}

\subsection{Properties of the representation}
\label{app:graph_properties}

\paragraph{Proposition 1 (Source-preserving paths).}
\label{prop:source_preservation}
Every directed path $(v_0,\ldots,v_m)$ in the source-visit graph $H_x$
has the form
\begin{equation}
 v_\ell=(r,j+\ell),\qquad \ell=0,\ldots,m,
 \label{eq:source_consistent_path}
\end{equation}
for one source $r$. If all visits on the path are indexable, its projection is a walk
in $G_x$; the converse need not hold.

\emph{Proof.} Every temporal edge preserves the source identifier and
increments the recorded position by one, proving the first statement by
induction. Projection maps each edge with indexable endpoints to $G_x$.
For the converse, sources $A\to X\to F$ and $B\to X\to G$ induce a
quotient walk $A\to X\to G$ without a single-source lift. \textit{QED.}
This proposition concerns the provenance of recorded experience: visit
identities retain which history and outcome belong to each source. It does
not rule out a valid newly composed cross-source path. Such a proposal
would require its own transition and information conditions; GC-OPD
instead supplies complete records as context for teacher scoring. Matching
locators do not establish that student and source possess identical
observations or preparations.

\paragraph{Proposition 2 (Monotonicity of recorded support).}
\label{prop:support_monotonicity}
Write $\mathcal A^+(z;\mathcal D)$ for the candidates in
Eq.~\ref{eq:recorded_support} at locator $z$ in library $\mathcal D$.
Fix the query locator, validity tests, and source costs. For nested libraries
$\mathcal D\subseteq\mathcal D'$ that preserve existing records and labels,
and with eligibility defined per source rather than by rank, let
\[
 m(z;\mathcal D)=\min_{(r,j)\in\mathcal A^+(z;\mathcal D)}d(r,j),
 \qquad\min\varnothing=+\infty.
\]
Then
\begin{equation}
\mathcal A^+(z;\mathcal D)\subseteq\mathcal A^+(z;\mathcal D'),
\qquad m(z;\mathcal D')\leq m(z;\mathcal D).
\label{eq:recorded_support_monotonicity}
\end{equation}
\emph{Proof.} Every existing candidate retains its locator and validity;
minimizing unchanged costs over a superset cannot increase the minimum.
\textit{QED.} This concerns available evidence before selection and
rendering, not student policy improvement, and does not compare
independently generated, nonnested libraries.

\paragraph{Representation size.}
For $K$ sources with at most $T$ committed decisions each,
$|\mathcal W_x|\leq K(T+1)$ and $|\mathcal F_x|\leq KT$; projection cannot
increase either count. Here $T$ bounds source length, including augmented
records, rather than the student evaluation horizon. Noncommitted attempts
and replay metadata are accounted for separately.

\subsection{Evidence-induced changes in the token signal}
\label{app:evidence_shift}
For the same student parameters, input history, and response tokens,
the student term in Eq.~\ref{eq:conditioned_signal} cancels between
two teacher contexts:
\begin{equation}
A^R_{t,i}-A^{R_0}_{t,i}
=\operatorname{sg}\!\left[\log q^R_{t,i}-\log q^{R_0}_{t,i}\right]
=\operatorname{sg}\!\left[\log\frac{q^R_{t,i}}{q^{R_0}_{t,i}}\right].
\label{eq:evidence_shift}
\end{equation}
This identity relates the scores in the \hyperref[ex:scoring_case]{case study} to the training signal;
it does not equate a probability shift with correctness.

\subsection{Local sensitivity of the native objective}
\label{app:gradient}
Use the episode-normalized weights from Appendix~\ref{app:training_configuration}.
Fix the sampled responses, input histories, teacher evidence, masks, and
rollout probabilities. Here $i$ indexes an unmasked response token,
including its episode and turn. At a common parameter point $\theta_0$,
suppose $\rho_i(\theta_0)=1$ and the OPD surrogate clipping is inactive.
Define
\[
g_i=\left.\nabla_\theta\log p_{\theta,i}\right|_{\theta=\theta_0},
\qquad \Delta_i=\log q_i^{R'}-\log q_i^R.
\]
Differentiating the detached surrogate in Eq.~\ref{eq:opd} gives
\[
\nabla\mathcal L_{R'}(\theta_0)-\nabla\mathcal L_R(\theta_0)
=-\sum_i w_i\Delta_i g_i.
\]
For one plain SGD step, let $\theta_R^+=\theta_0-\eta\nabla\mathcal L_R(\theta_0)$
and define the change at a fixed-context probe token $j$ by
$\delta_R\log p_j=\log p_{\theta_R^+,j}-\log p_{\theta_0,j}$. Then
\[
\delta_{R'}\log p_j-\delta_R\log p_j
=\eta\sum_iw_i\Delta_i\langle g_j,g_i\rangle+O(\eta^2).
\]
For the mean log probability of fixed action-span tokens, replace $g_j$
by their mean gradient. Self-token and cross-token terms can oppose each
other. This local sensitivity identity excludes AdamW preconditioning,
gradient-norm clipping, active OPD clipping, and subsequent changes in
sampled trajectories.

\section{Computational cost details}
\label{app:compute_details}

\begin{table}[!tbp]
\centering\small
\caption{\textbf{ScienceWorld preparation and student-training costs.} Times are measured separately for each stage. Student-training rows use a 1.7B student, a 32B teacher, and two epochs. GC-OPD includes one vanilla OPD epoch and one graph-conditioned epoch.}
\label{tab:computation_context}
\setlength{\tabcolsep}{4pt}
\begin{tabularx}{\linewidth}{@{}Xrrr@{}}
\toprule
Stage / method & H20 GPUs & Time (h) & GPU-h\\
\midrule
$K=16$ collection & 64 & 3.32 & 212.48\\
GRPO teacher training & 64 & 18.27 & 1,169.28\\
\midrule
Vanilla OPD & 8 & 5.91 & 47.31\\
TCOD-B2F & 8 & 5.70 & 45.58\\
TCOD-F2B & 8 & 3.74 & 29.94\\
FTB & 8 & 7.27 & 58.18\\
TurnOPD & 8 & 4.35 & 34.81\\
GC-OPD & 8 & 8.60 & 68.79\\
\bottomrule
\end{tabularx}
\end{table}

\paragraph{Student training.}
The two-epoch student-training runs use eight H20 GPUs. Vanilla OPD takes
5.91 hours; GC-OPD takes 8.60 hours for one vanilla OPD epoch followed by
one graph-conditioned epoch. The corresponding costs are 47.31 and
68.79 GPU-hours. For TCOD-B2F, TCOD-F2B, FTB, and TurnOPD, the reported
times sum the complete 190-update training logs, including checkpoint
saving. Times and GPU-hours are rounded for display.

\paragraph{Teacher preparation.}
Collecting the $K=16$ execution library takes 3.32 hours on 64 H20 GPUs,
or 212.48 GPU-hours. GRPO teacher training takes 18.27 hours on 64 H20
GPUs, or 1,169.28 GPU-hours, using sampling groups of 16. These preparation
times are measured separately from student training.
Figure~\ref{fig:homepage_teaser}(b) sums each preparation stage with the
corresponding student-training cost.

\end{document}